\documentclass[letterpaper]{article} % DO NOT CHANGE THIS
\usepackage{aaai25}  % DO NOT CHANGE THIS
\usepackage{times}  % DO NOT CHANGE THIS
\usepackage{helvet}  % DO NOT CHANGE THIS
\usepackage{courier}  % DO NOT CHANGE THIS
\usepackage[hyphens]{url}  % DO NOT CHANGE THIS
\usepackage{graphicx} % DO NOT CHANGE THIS
\usepackage{natbib}  % DO NOT CHANGE THIS AND DO NOT ADD ANY OPTIONS TO IT
\usepackage{caption} % DO NOT CHANGE THIS AND DO NOT ADD ANY OPTIONS TO IT
\usepackage{algorithm}
\usepackage{algorithmic}
\usepackage{amsmath}
\usepackage{blkarray}
\usepackage{subcaption}
\usepackage{multirow}
\usepackage{newfloat}
\usepackage{listings}
\DeclareCaptionStyle{ruled}{labelfont=normalfont,labelsep=colon,strut=off} % DO NOT CHANGE THIS
\floatstyle{ruled}
\newfloat{listing}{tb}{lst}{}
\floatname{listing}{Listing}
\title{Process-aware Human Activity Recognition}
\author{
    Jiawei Zheng, Petros Papapanagiotou, Jacques D.\ Fleuriot, Jane Hillston
}
\affiliations{
    School of Informatics,
    University of Edinburgh\\
    jw.zheng@ed.ac.uk, petrospapapan@gmail.com, jdf@ed.ac.uk, jane.hillston@ed.ac.uk
}

\usepackage{bibentry}
\begin{document}

\maketitle

\begin{abstract}

Humans naturally follow distinct patterns when conducting their daily activities, which are driven by established practices and processes, such as production workflows, social norms and daily routines. Human activity recognition (HAR) algorithms usually use neural networks or machine learning techniques to analyse inherent relationships within the data. However, these approaches often overlook the contextual information in which the data are generated, potentially limiting their effectiveness. We propose a novel approach that incorporates process information from context to enhance the HAR performance. Specifically, we align probabilistic events generated by machine learning models with process models derived from contextual information. This alignment adaptively weighs these two sources of information to optimise HAR accuracy. Our experiments demonstrate that our approach achieves better accuracy and Macro F1-score compared to baseline models.

\end{abstract}

\section{Introduction}

Analysing patterns of human activities offers a multitude of benefits across various domains, including interactive gaming, fitness activity monitoring, assisted living and healthcare monitoring~\cite{guptaHumanActivityRecognition2022,liuOverviewHumanActivity2022}. For example, in healthcare, monitoring stroke patients' daily activities can track their recovery progress and lead to interventions when necessary. With the rapid development of sensor technologies and smart devices, such as smartphones and cameras, large amounts of data related to human activities and behaviour are generated. Human activity recognition (HAR) techniques identify and recognise human activities based on the generated data from these sources. 

However, core HAR methods mainly focus on leveraging the power of machine learning (ML) and deep learning (DL) techniques to extract information from data for activity recognition~\cite{guptaHumanActivityRecognition2022}. These HAR methods relies solely on analysing the inherent traits and relationships within the data. Such an approach can only learn from whatever data are fed to it, but neglect the context in which data are generated. This may limit their effectiveness~\cite{guptaRoadmapDomainKnowledge2020}, resulting in low accuracy due to inter-class similarity. For example, the activities of \textit{drinking from a cup} and \textit{answering a phone} involve similar arm movements~\cite{guptaObjectsActionApproach2007}. Thus, it is difficult to distinguish these similar activities based on motion data alone.

To address this limitation, the idea of incorporating domain knowledge or contextual information into ML and DL methods has been investigated to improve the accuracy and interpretability~\cite{guoDiCNNDomainKnowledgeInformedConvolutional2023a}. Humans inherently follow distinct patterns in their daily activities, influenced by established best practices and workflows. These include production processes, social norms and habitual routines, such as adhering to a recipe when preparing a meal or following a typical daily schedule. 
Incorporating this type of information into the traditional data-driven HAR models mentioned above can help enhance their performance. Taking the example of distinguishing between the activities of \textit{drinking from a cup} and \textit{answering a phone}, it is helpful to consider the typical sequence of actions associated with each activity. For instance, the process of drinking often starts with approaching the location of the cup, followed by reaching for and picking up the cup, then drinking from it and finally putting down the cup. By understanding and recognising this process, we can more accurately infer that if a series of actions match this pattern, the current activity is likely \textit{drinking from a cup} rather than \textit{answering a phone}.

Moreover, capturing this process information poses challenges in 
domains where activities may not follow a strictly defined sequence. Across domains, the variability of how tasks are performed makes it difficult to define a structured process. For instance, the same activity of drinking from a cup could involve different sequences, e.g., an individual could be interrupted when they receive a text message or doorbell ringing, altering the usual sequence.

Our work aims to enhance the performance of traditional HAR models by incorporating process information from context. In this paper, we propose a novel approach based on alignment conformance checking algorithm~\cite{carmonaConformanceCheckingRelating2018,zhengAlignmentbasedConformanceChecking2022HICSS}, which aligns a process model with an event log. The proposed approach learns from both the outputs of HAR models and processes information from domains. It can adaptively weights these two sources of information during the training phase to achieve the best HAR accuracy. An experimental study is investigated based on a dataset collected by cameras whilst observing people eat~\cite{razaEatSenseHumanCentric2023}. The result shows that our approach can achieve better accuracy and Macro F1-score compared to two well-known Graph Convolutional Network algorithms for activity recognition.

\section{Motivating example} \label{sec:motivating example}

Eating behaviour monitoring has seen increased attention in recent research, particularly in the context of social care and healthcare~\cite{tufanoCapturingEatingBehavior2022,razaEatSenseHumanCentric2023}. Recognising an individual's actions such as picking up food and placing it in their mouth, can help analyse their eating behaviour and provide valuable insights into their health condition. For example, consistently monitoring how quickly a person eats, the size of the bites they take, or their ability to steadily hold utensils can be indicators of potential health issues, such as neurological disorders or physical impairments~\cite{tufanoCapturingEatingBehavior2022}. Moreover, the importance of recognising eating actions also extends to the development of assistive technologies, such as robotic arms for help with eating, e.g., for individuals with disabilities or the elderly. 

Existing work has investigated how ML and DL-based classification models can be used for recognising activities~\cite{wuAIAssistedFoodIntake2022,razaEatSenseHumanCentric2023}. These models are proficient in extracting features from various types of input data, such as images and sensor readings. They generate an output matrix that represents the prediction probabilities for the input across different classes of eating actions, like \textit{chewing}, \textit{picking up cutlery}, and \textit{putting down cutlery}. This output matrix is a probability distribution across possible activity classes for any given input. The recognised activity is determined to be the one associated with the highest probability in the matrix.

However, due to the similarities between different classes of actions, selecting the class with the highest probability does not always guarantee accuracy. This is because similar actions may result in similar probabilities. For example, the algorithm may classify some sensor input associated with the activity of \textit{putting down cutlery} as the \textit{picking up cutlery} activity with 51\% probability and \textit{putting down cutlery} with 49\% probability, resulting in misclassification.

Eating behaviour possesses inherent process information embedded within the execution of eating activities. For example, there is a typical order of activities, such as \textit{picking up a fork}, \textit{picking up food using the fork}, \textit{putting food into mouth}, \textit{chewing}, etc. Understanding these aspects, such as the processes involved in eating, allows us to capture the context in which the activity is performed, thus improving our ability to make accurate activity recognition. For example, \textit{picking up cutlery} is likely to be followed by an activity related to food consumption (like \textit{picking up food using a fork}), whereas \textit{putting down cutlery} might be followed by activities like wiping one's mouth or reaching for a drink. Therefore, by incorporating the process information, we can reduce the uncertainty of activity recognition algorithms caused by inter-class similarities and enhance the accuracy of activity recognition.

However, eating behaviour also includes a degree of flexibility in the sequence and manner of performing eating actions, which complicates the task of establishing a structured process that could be incorporated to enhance activity recognition. For example, while a typical sequence at mealtime might involve picking up cutlery, using the cutlery to pick up food, and then eating the food, this sequence can be altered based on situational factors. For instance, a person could be interrupted by a phone call or need to attend to a child, causing them to put down their cutlery unexpectedly and alter their usual sequence.

Moreover, individuals often have personal eating habits that vary widely, such as the order in which they eat their food. Some might prefer starting with lighter items like soup before progressing to heavier main dishes, while others might dive straight into the main course. Additionally, the type of food being consumed often determines the utensils required. For example, soup is eaten with a spoon, steak requires a knife and fork, and sushi is typically eaten with chopsticks. These variations mean that eating different types of food can involve distinct sequences of actions. Sometimes, people might choose to eat with their hands instead of using cutlery, bypassing some of the steps in the typical sequence. These variations make it challenging to capture structured process information of eating behaviour. 

\section{Preliminary}

Our approach is built upon an alignment-based conformance checking algorithm for uncertain data, proposed by Zheng et al., known as \texttt{ProbCost}~\cite{zhengAlignmentbasedConformanceChecking2022HICSS}. Next, we provide a brief description of relevant concepts so that this paper is self-contained.

\begin{figure}[]
    \centering%
    \includegraphics[width=.85\columnwidth]{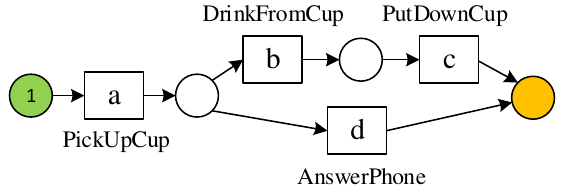}
    \caption{Example of a process model.}%
    \label{fig:process model}%
\end{figure}

\texttt{ProbCost} aligns probabilistic event data with process information and needs two components to achieve this. The first is a \textit{process model}, which represents the process information. An example process model, related to the drinking scenario is depicted in Figure~\ref{fig:process model}. The second component is a sequence of probabilistic events, each corresponding to one of the possible activities. For instance, consider a log of 3 events: $e_{0}, e_{1}, e_{2}$, as modelled in~\eqref{eq:example events}. The event $e_{0}$ has a $0.6$ probability of corresponding to \textit{PickUpCup} activity and $0.4$ probability of corresponding to \textit{PutDownCup} activity.
\begin{small}
\begin{equation}
    \mathbf{P} = 
        \begin{blockarray}{c@{\hspace{1pt}}ccc@{\hspace{3pt}}}
         & e_{0}   & e_{1} & e_{2}  \\
        \begin{block}{c@{\hspace{5pt}}|@{\hspace{2pt}}
    |@{\hspace{2pt}}ccc@{\hspace{2pt}}|@{\hspace{2pt}}|}
        PickUpCup & 0.6 & 0 & 0.4\\
        DrinkFromCup & 0 & 0.3 & 0\\
        PutDownCup & 0.4 & 0 & 0.6\\
        AnswerPhone & 0 & 0.7 & 0\\
        \end{block}
    \end{blockarray}
    \label{eq:example events}
\end{equation}
\end{small}

\texttt{ProbCost} combines these two components as a synchronous product net, which involves three types of move: \textit{synchronous moves (SM)}, \textit{model moves (MM)} and \textit{log moves (LM)}. A \textit{SM} indicates that the event in the log and the activity in the process correspond to each other. A \textit{MM} indicates a misalignment where an activity should have been executed according to the process model, but there is no corresponding event in the log. A \textit{LM} indicates that an activity has been executed, which was unexpected according to the process model. The principle of the alignment tries to find an optimal movement path between the process model and the event log using a cost function (\ref{eq:cost_function}) for each type of move. Given this, the problem of the alignment is reduced to searching for a movement path with the minimum cost using the A* algorithm~\cite{russell2016artificial}.
\begin{small}
\begin{equation}
c(t) = \begin{cases} 
% 0, & t \in MM in hidden acti \\
-\log(w(t)), & t \in \textit{SM} \\
-\log(w(t)) - \log(\epsilon), & t \in \textit{LM} \\
-\log(\epsilon), & t \in \textit{MM} \\
\end{cases}
\label{eq:cost_function}
\end{equation}
\end{small}

In the cost function \eqref{eq:cost_function}, $w(t)$ represents the associated probability of an activity for a given event. For example, consider the events in \eqref{eq:example events}, $w(PickUpCup)$ for $e_{0}$ is 0.6. There is also a parameter $\epsilon \in (0, 1)$. A lower value of $\epsilon$ means that the alignment allows events with lower probability to achieve a \textit{SM} with a matching activity in the process model. In contrast, a higher value of $\epsilon$ requires the event to have a higher probability to align with the activity in the process model as a \textit{SM}, following the principle of choosing the most probable activity. The intuition behind this is to weigh the trust between the process model and the sequence of probabilistic events to achieve the alignment. Specifically, a lower value of $\epsilon$ places more weight on the process model, accepting lower probability activities, while a higher value places more weight on probabilistic events, choosing higher probability activities.

Consider once more the sequence of probabilistic events represented in \eqref{eq:example events} and also the process model in Figure~\ref{fig:process model}, \texttt{ProbCost} can obtain different alignment by adjusting the values of $\epsilon$ to match the actual sequence of events. For example, if the actual activities associated with each event are \textit{PickUpCup}, \textit{DrinkFromCup} and \textit{PutDownCup}, respectively, a lower threshold of $\epsilon$ (0.4) can result in a perfect alignment (Table~\ref{tab:example alignment 0.4}) with process model by choosing the activity \textit{DrinkFromCup} for the second event, as it matches to the process model and results the minimum cost of movement, even though \textit{DrinkFromCup} has lower probability compared to \textit{AnswerPhone}. Conversely, if the actual sequence of events is \textit{PickUpCup}, \textit{AnswerPhone} and \textit{PutDownCup}, as indicated by the most probable classes, a high value of $\epsilon$ (0.9) can select the most probable activity, i.e., \textit{AnswerPhone}, to achieve the alignment (Table~\ref{tab:example alignment 0.9}).

\begin{table}[h]
    \centering
    \begin{subtable}{\columnwidth}
    \centering
    \resizebox{\columnwidth}{!}{
    \begin{tabular}{c|c|c|c|}
    \small{Event log}     & \small{PickUpCup} & \small{DrinkFromCup} & \small{PutDownCup}  \\ \hline
    \small{Process model} & \small{PickUpCup} & \small{DrinkFromCup} & \small{PutDownCup} 
    \end{tabular}
    }
    \caption{The alignment result with $\epsilon = 0.4$.}
    \label{tab:example alignment 0.4}
    \end{subtable}
    
    \begin{subtable}{\columnwidth}
    \centering
    \resizebox{\columnwidth}{!}{
    \begin{tabular}{c|c|c|c|}
    \small{Event log}     & \small{PickUpCup} & \small{AnswerPhone} & \small{PutDownCup}   \\ \hline
    \small{Process model} & \small{PickUpCup} & \small{AnswerPhone} & $\gg$  
    \end{tabular}
    }
    \caption{The alignment result with $\epsilon=0.9$.}
    \label{tab:example alignment 0.9}
    \end{subtable}
    \caption{The alignment results with different $\epsilon$.}
    \label{tab:example alignment}
\end{table}

\section{Method}\label{sec:method of HAR}

Our proposed approach for process-aware human activity recognition consists of 4 stages: i) extracting a probability distribution matrix from ML or DL models, ii) discovering process models, iii) conducting an alignment by \texttt{ProbCost}, and iv) retrieving activities from the alignment. 

\begin{figure*}
    \centering
    \includegraphics[width=0.8\linewidth]{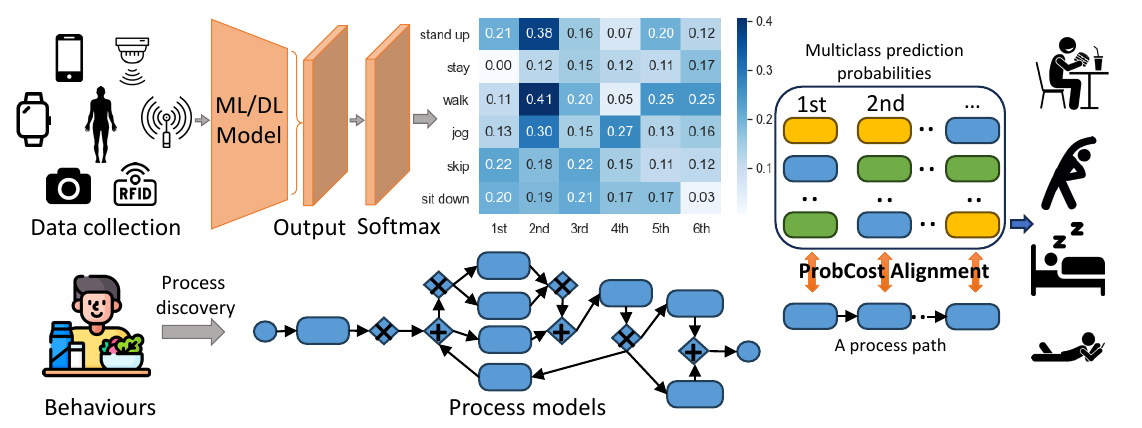}
    \caption{The framework of process-driven human activity recognition.}
    \label{fig:framework}
\end{figure*}

Figure~\ref{fig:framework} shows our framework, which takes the output of multiclass prediction probabilities from ML or DL algorithms as probabilistic events. Process models are discovered from activity-labelled data. Taking both probabilistic events and process models as input, we apply \texttt{ProbCost} to achieve an alignment. \texttt{ProbCost} is used to calibrate the results of ML/DL-based activity recognition algorithms by incorporating process information. An optimal alignment is obtained between the process model and the probabilistic event log. Then we retrieve the events from the achieved alignment as the final recognised activities. We present the details of different parts in the framework next.

\subsection{Extracting probabilistic traces}

Traditional ML and DL-based classification models can provide a probability distribution over a set of potential classes for a given input, termed as \textit{multiclass prediction probabilities}. For example, neural networks often conclude their prediction output with a softmax layer, while decision trees and random forest algorithms can estimate probabilities for each class~\cite{JamesIntroductionML2023}. These probabilistic outputs serve as a measure of the model's confidence level in its prediction. Each class's probability reflects the likelihood that the given input belongs to that class. 

The probabilities associated with each potential activity can be extracted as probabilistic events. We construct a \textit{probabilistic event trace} by following the chronological order in which activities occur, which is modelled as a matrix~\eqref{eq:multiclass distribution}.
\begin{small}
\begin{equation}
\label{eq:multiclass distribution}
% $
% \resizebox{.9\columnwidth}{!}
% {
\scalebox{0.8}{
$\begin{blockarray}{c@{\hspace{1pt}}ccccc@{\hspace{3pt}}}
         & 1st & 2nd & ... & (m-2)th & (m-1)th \\
        \begin{block}{c@{\hspace{5pt}}|@{\hspace{2pt}}
    |@{\hspace{2pt}}ccccc@{\hspace{2pt}}|@{\hspace{2pt}}|}
  a_{0} & p_{0,0} & p_{1,0} & ... & p_{m-2,0} & p_{m-1,0} \\
  a_{1} & p_{0,1} & p_{1,1} & ... & p_{m-2,1} & p_{m-1,1} \\
  . & . & . & ... & . & . \\
  . & . & . & ... & . & . \\
  a_{n-2} & p_{0,n-2} & p_{1,n-2} & ... & p_{m-2,n-2} & p_{m-1,n-2} \\
  a_{n-1} & p_{0,n-1} & p_{1,n-1} & ... & p_{m-2,n-1} & p_{m-1,n-1} \\
\end{block}
\end{blockarray}$
% }
}
\end{equation}
\end{small}

Each row in the matrix corresponds to a potential class of activity for which the model provides a probability. Each column in the matrix represents a specific activity or time step at which predictions are made. Each cell in the matrix contains the probability that the model assigns to the corresponding class for the given activity. The sum of the probabilities for each activity should be $1$.

\subsection{Process discovery}

We use process mining techniques~\cite{vanderaalstProcessMining2016} to discover process models in human behaviours based on data collected from various sensors. This data labelled with specific activities is used to build an event log, which contains the relevant information for process mining, including timestamps, performed activities, and cases. A case can be defined based on a specific period, or by specific instances of an activity, such as each mealtime. Then the event log can be used in process mining tools for discovering processes, such as PM4Py~\cite{PM4PYpaper2019}.

There are various process discovery algorithms, each suited to different scenarios. The most commonly used algorithms are \textit{inductive miner}~\cite{leemans2014process} and \textit{heuristic miner}~\cite{weijtersProcessMiningHeuristicsMiner2006}. Inductive miner is known for its robustness in creating process models that guarantee logs can be replayed. The Inductive miner is particularly effective in structured processes, such as manufacturing, where activities follow strict workflows. Heuristic miner is particularly effective for discovering meaningful process models in scenarios where data exhibits considerable variation yet contains structured patterns. It is well-suited for such settings because it focuses on identifying the most frequent paths in the event log, capturing the main activity flows. Compared to the Inductive miner which might struggle with flexibility and irregularities, the Heuristic miner emphasis on frequency helps to identify the main behaviour in the event log.

\subsection{Alignment}

With the discovered process models and extracted probabilistic event traces, we can perform alignment between these two components to calibrate the classification results. The aim of the alignment is to find a possible activity class in each probabilistic event by incorporating inherent structured process information to enhance activity recognition. 

We apply \texttt{ProbCost} to achieve alignments between the probabilistic event traces output from ML predictions with the process model. In \texttt{ProbCost}, the probabilistic event trace and the discovered process model are combined as a synchronous product net, where each movement is associated with a cost defined by the cost function (see~\eqref{eq:cost_function}). The output of \texttt{ProbCost} is an alignment (movement path) with minimal cost. Activity classes selected in the achieved alignment serve as the recognised activities.

The parameter $\epsilon$ adjusts the cost associated with each type of move in the alignment based on the likelihood that a particular activity occurred as predicted by ML algorithms. This allows the alignment to take into account probabilistic events that may have a lower probability but align better with the process model. The threshold $\epsilon$ serves as a control mechanism to balance the trust between these two aspects, i.e., the probabilistic events and the process model. 

We treat $\epsilon$ as a hyperparameter to tune the cost function with the aim of achieving the highest activity recognition accuracy. To determine the optimal setting of $\epsilon$, we utilise a portion of our datasets as a validation set for tuning and selecting the $\epsilon$. By adjusting the value of $\epsilon$ and observing the resulting changes in accuracy on the validation set, we select the $\epsilon$ that can achieve the highest accuracy in the validation set as the optimal value of $\epsilon$ for final evaluation.

\section{Experimental study}\label{sec:experiment HAR}

\subsection{Dataset description}

Our experimental study is based on the EatSense dataset, which uses RGBD cameras to capture the eating behaviours of participants~\cite{razaEatSenseHumanCentric2023}. This dataset records the upper-body movements of individuals while they eat. It encompasses data from participants who use a range of utensils to consume different types of food. Within this dataset, 16 distinct eating actions such as \textit{chewing}, \textit{drinking}, \textit{food in hand at table} have been densely labelled. The full list of actions is shown in Table~\ref{tab:eating actions appendix} in the Appendix.

The dataset comprises a trimmed component, where the videos have already been manually segmented into separate clips for each action. These trimmed clips are arranged in chronological sequence, mirroring the order in which the actions were performed by the individuals. Each video is a complete eating behaviour process for one subject. 
The dataset includes videos capturing the eating behaviours of six subjects eating nine different types of food, such as rice, toast, roti, etc. The distribution of videos across these subjects and food types is detailed in Table~\ref{tab:number of videos appendix} in the Appendix.

For each video, the skeleton of upper-body poses is extracted. The datasets consider 8 human body keypoints, including the nose, chest, right and left shoulder, etc. The extracted skeleton data can be used by Graph Convolutional Networks (GCNs) for activity recognition~\cite{shiTwoStreamAdaptiveGraph2019}.

\subsection{Experimental design}
 
Recognising that different individuals consuming different types of food entails a distinct set of processes, we undertake a classification of the videos into distinct categories through two strategic approaches to accurately capture the process information in eating behaviour. Firstly, we focus on discovering the process model associated with each individual, recognising the personal patterns and habits in their eating behaviours. The second strategy is to discover the process models linked to consuming various types of food, taking into account that each food category demands specific eating actions, sequences, and possibly different utensils. 

Within the EatSense dataset, the number of videos associated with \textit{subject 1} is over 10, which although it represents a relatively small dataset does allow for meaningful analysis. Similarly, there are 13 videos related to \textit{eating roti}. 

We train machine learning models and carry out a detailed exploration of how various actions are coordinated in the process of eating, focusing our analysis on the videos related to \textit{subject 1} and the videos capturing \textit{eating roti}.

To get probabilistic events, we apply GCNs for activity recognition based on skeleton data. GCNs are widely used for skeleton-based action recognition and have achieved remarkable performance since they are adept at representing human body skeletons as spatio-temporal graphs, enabling the modelling of complex movements and interactions among different body parts over time~\cite{yanSpatialTemporalGraph2018}. We use the top two performing GCNs models as identified in the study by~\citeauthor{razaEatSenseHumanCentric2023} as our baselines. These two models are two-stream Adaptive Graph Convolutional Network (2s-AGCN)~\cite{shiTwoStreamAdaptiveGraph2019} and Channel-wise Topology Refinement Graph Convolutional Network (CTR-GCN)~\cite{chenChannelwiseTopologyRefinement2021}. Our aim is to use these GCN models as baselines for recognising eating activities and to extract probabilistic event traces from their outputs.

In each category, we choose 60\% of the videos as our training set, 20\% of the videos as the validation set and the rest of the videos as the testing set. The training set is used for the GCNs models and to discover the process model. We use the validation set to tune the parameter of process discovery and the threshold $\epsilon$ in the \texttt{ProbCost} algorithm.

We use the 3D keypoints skeleton of 8 upper-body joints from EatSense datasets as the input for the GCNs models. The GCNs models are trained using the training set, and its performance is evaluated on the testing set. Subsequently, we convert the output predictions of the GCNs models into a probabilistic matrix by applying a softmax function to the results of their output layer. 

With the discovered process model and extracted probabilistic event traces, we apply \texttt{ProbCost} to achieve an alignment between them. The confidence threshold $\epsilon$ is tuned on the validation set with the aim of maximising the activity recognition accuracy. The tuned value of $\epsilon$ is applied to the testing set for the performance evaluation. 

Subsequently, we compare the activity recognition performance, contrasting the results obtained from the original GCNs models against those achieved after applying the \texttt{ProbCost} alignment. We use accuracy and macro F1-score to evaluate the performance. Accuracy provides a straightforward measure of the model's overall correct predictions, while the macro F1-score offers a balanced view of the precision and recall across all classes.

\subsection{Process discovery}

As we mentioned in Section~\ref{sec:motivating example}, eating behaviour allows a degree of flexibility in performing the associated actions. Therefore, we employed the Heuristic miner (implemented in PM4Py) to discover the processes associated with \textit{subject 1} and \textit{eating roti}. The Heuristic miner requires setting a dependency threshold to eliminate less frequent paths between activities. We varied the dependency threshold from 0.8 to 0.95 with a 0.05 step for discovering different process models based on the training set. We compute fitness and precision metrics ~\cite{vanderaalstProcessMining2016} of the process models using PM4Py. The fitness measures how well the discovered process model can replay the observed behaviours. Precision essentially measures the model's ability to generate only those traces that are actually present in the event logs, thereby evaluating how closely the process model aligns with the real-world processes it aims to represent. We also measure the F-score of the precision and fitness by $2 \times \frac{\textit{Precision} \times \textit{Fitness}}{\textit{Precision} + \textit{Fitness}}$. 
The result of discovered process model for \textit{subject 1} is shown in Table~\ref{tab:fitness process model}, with the highest F-score achieved at a dependency threshold of 0.85.

\begin{figure*}
    \centering
    \begin{subfigure}[t]{0.45\textwidth}
        \centering
        \includegraphics[width = \textwidth]{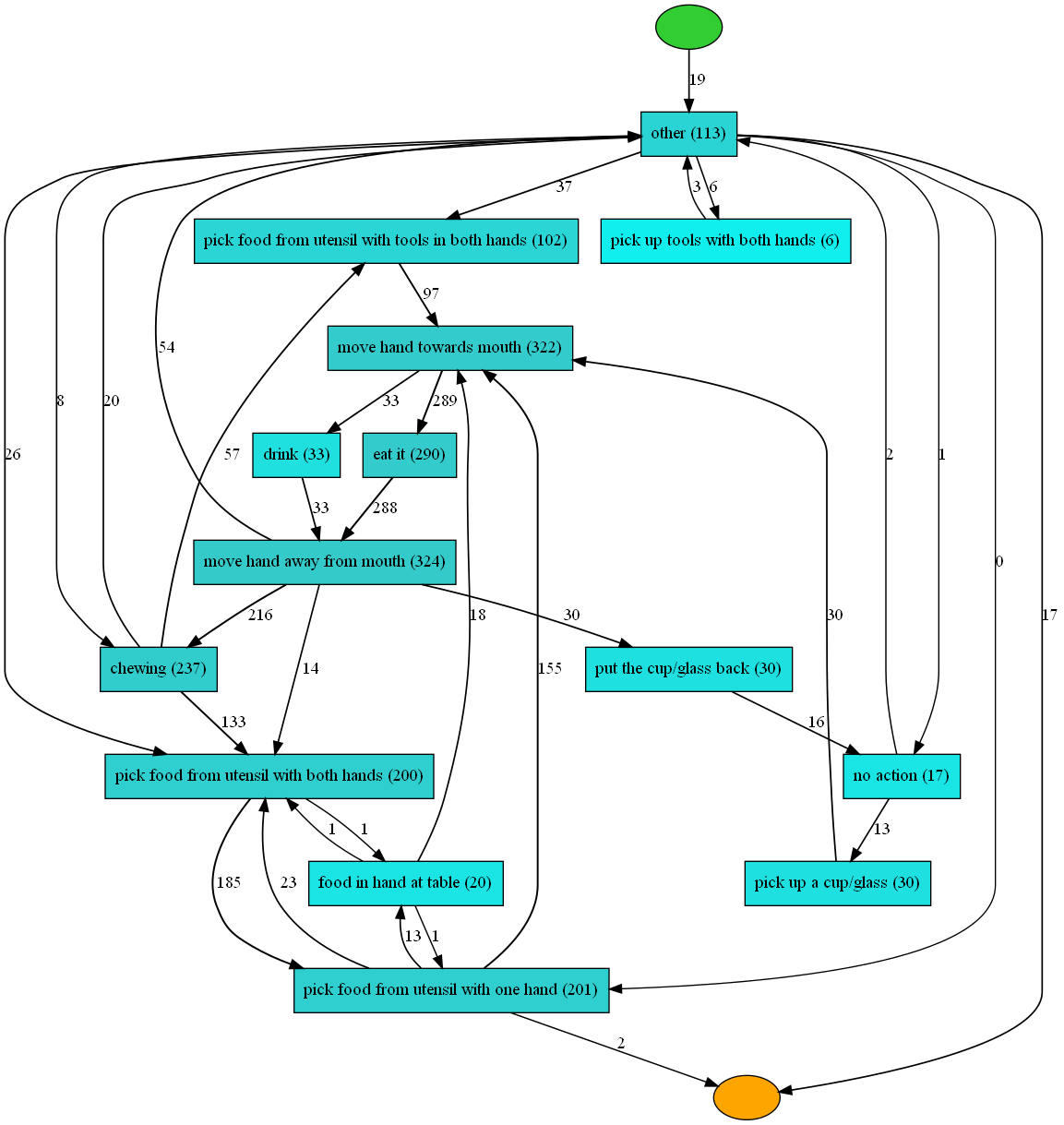}
        \caption{Process model of \textit{subject 1} with dependency threshold 0.85.}
        \label{fig:sb1Process}
    \end{subfigure}%
    \qquad
    \begin{subfigure}[t]{0.45\textwidth}
        \centering
        \includegraphics[width = \columnwidth]{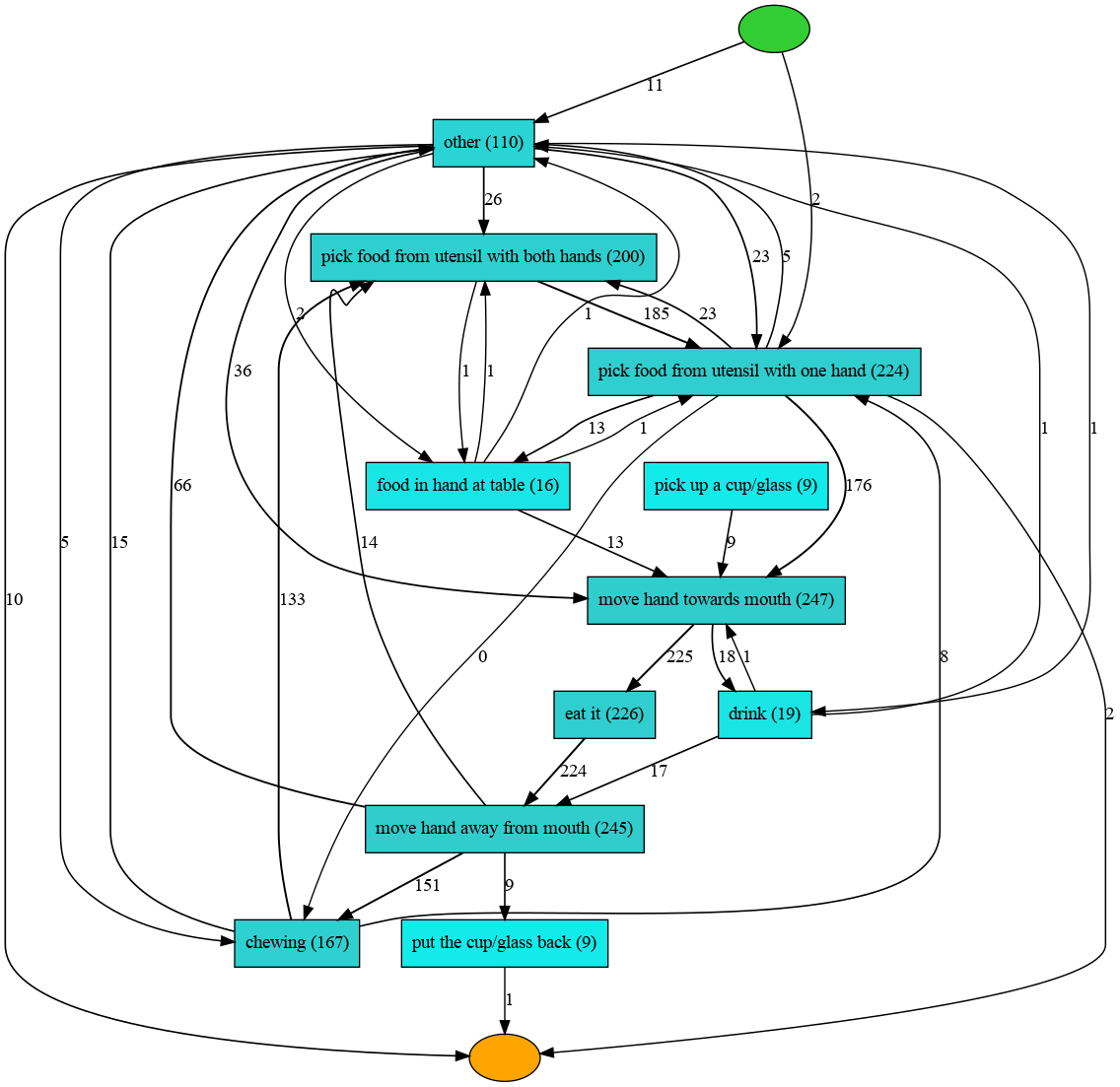}
        \caption{Process model of \textit{eating roti} with dependency threshold 0.9.}
        \label{fig:RotiProcess}
    \end{subfigure}%

    \caption{Discovered process models. The number in the node label indicates the frequency of the corresponding activity. The number in the arc indicates the frequency of the corresponding path.}
    \label{fig:discovered process model}
\end{figure*}

The corresponding process model of \textit{subject 1} is shown in Figure~\ref{fig:sb1Process}, where the numbers indicate the frequency of relevant activities and paths. It shows that the eating process of \textit{subject 1} starts from \textit{other} activity, which represents some preparatory activities, such as adjusting seating and setting the table, involved in the eating but not specifically catalogued within our dataset. Then \textit{subject 1} tends to perform the action of \textit{pick food from utensil with tools in both hands} or \textit{pick up tools with both hands}, demonstrating the use of cutlery or other eating instruments. This action is closely followed by the action of \textit{move hand towards month}. Once the food is brought to the mouth, the primary activity is \textit{eating it}, where the food is consumed. Occasionally, this step alternates with \textit{drink}, indicating the consumption of a beverage as part of the meal. After each instance of eating or drinking, \textit{subject 1} performs the action of \textit{move hand away from mouth}. Then the subject performs \textit{chewing}, where the food is processed for swallowing, \textit{put the cup/glass back} if \textit{drink} is performed before, or picking food again for eating either with tools or without tools depending on the consumed food. 

We can see the most frequent path of the eating process is starting from \textit{other}, then \textit{pick food from utensil with tools in both hands}, \textit{move hand towards month}, \textit{eat it}, \textit{move hand away from month}, \textit{chewing}, etc. It shows that this process model effectively captures the structured action sequences of eating. Thus, we choose the process model of \textit{subject 1} discovered at the dependency threshold 0.85 for alignment. 

\begin{table}[t]
\centering
% \begin{small}
\resizebox{\columnwidth}{!}{
\begin{tabular}{ccccc}
\hline
Category & \begin{tabular}[c]{@{}c@{}}Dependency\\  threshold \end{tabular} & Fitness & Precision & F-score \\ \hline
\multirow{4}{*}{subject 1} & 0.80 & 0.43 & 0.64 & 0.52 \\
 & 0.85 & 0.72 & 0.54 & \textbf{0.62}  \\ 
 & 0.90 & 0.76 & 0.31 & 0.44  \\ 
 & 0.95 & 0.41 & 0.41 & 0.41 \\  \hline

% \hline
% Dependency threshold & Fitness & Precision & F-score\\ \hline
\multirow{4}{*}{eating roti} & 0.80 & 0.61  & 0.30 & 0.40\\
& 0.85        &  0.64    & 0.34            &   0.44  \\ 
& 0.90        &  0.84            &    0.42            &   \textbf{0.56}  \\ 
& 0.95 & 0.81 & 0.42 & 0.55  \\ \hline
\end{tabular}
}
% \end{small}
\caption{The fitness, precision and F-score of \textit{subject 1} and \textit{eating roti} process model in different dependency threshold.}
\label{tab:fitness process model}
\end{table}

The evaluation of the discovered process model associated with \textit{eating roti} is shown in Table~\ref{tab:fitness process model}. It shows that the highest F-score is achieved at a dependency threshold of 0.9. We visualise the discovered process at 0.9 dependency threshold in Figure~\ref{fig:RotiProcess}. It depicts the process of eating roti, i.e., starting from \textit{pick food from utensil with both hands}, \textit{pick food from utensil with one hand}, \textit{move hand away from month}, \textit{chewing}, etc. Compared to the discovered process model of \textit{subject 1} (Figure~\ref{fig:sb1Process}), there is no \textit{pick food from utensil with tools in both hands}, because eating roti does not need tools. We adopt this process model of eating roti for subsequent alignment.

\subsection{Results and discussion}

\begin{figure}[h]
    \centering
    \begin{subfigure}[t]{\columnwidth}
        \centering
            \includegraphics[width=.7\columnwidth]{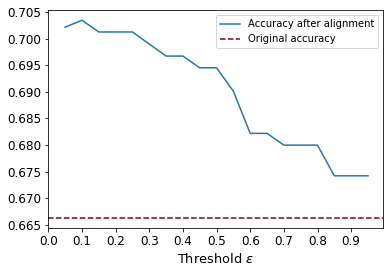}
            \caption{Accuracy across varying confidence thresholds $\epsilon$ for subject 1.}
            \label{fig:TuningThresholdVal}
        \end{subfigure}%
        
        \begin{subfigure}[t]{\columnwidth}
            \centering
            \includegraphics[width=.7\columnwidth]{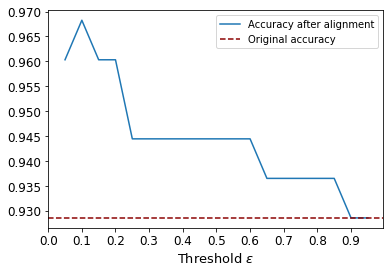}
            \caption{Accuracy across varying confidence thresholds $\epsilon$ for eating roti.}
            \label{fig:TuningThresholdValFood}
        \end{subfigure}%
    \caption{Accuracy of activity recognition across varying confidence thresholds $\epsilon$ in the validation set.}
    \label{fig:TuningThresholdValAll}
\end{figure}

Firstly, we focus on investigating activity recognition results for the data related to \textit{subject 1}. The confidence threshold $\epsilon$ in \texttt{ProbCost} is tuned in the validation set. We set $\epsilon$ from 0.05 to 1 with a 0.05 step to test the activity recognition accuracy. The average accuracy in validation set with different $\epsilon$ is shown in Figure~\ref{fig:TuningThresholdVal}. The result shows it achieves the highest accuracy when the confidence threshold $\epsilon$ equals $0.1$. Thus, we set the $\epsilon$ as $0.1$ for the evaluation on the testing set.

Table~\ref{tab:performance comparison} displays the activity recognition performance, comparing the original results from the 2s-AGCN and CTR-GCN models with their performance post-alignment with the process model via ProbCost. The result shows an improvement in both metrics for the 2s-AGCN and CTR-GCN methods. This improvement indicates the benefit of incorporating process information into the activity recognition.

\begin{table}[tb]
\centering
\resizebox{\columnwidth}{!}{
\begin{tabular}{cccccc}
\hline
Category & Methods & Accuracy & \begin{tabular}[c]{@{}c@{}}Macro\\  F1 \end{tabular} & \begin{tabular}[c]{@{}c@{}}\textbf{Accuracy}\\  \textbf{Alignment} \end{tabular} & \begin{tabular}[c]{@{}c@{}}\textbf{Macro F1}\\  \textbf{Alignment} \end{tabular} \\ \hline
\multirow{2}{*}{subject 1} & 2s-AGCN           & 0.67     & 0.44     & 0.70 & 0.46\\
& CTR-GCN &    0.57  &   0.32  & 0.71 & 0.48 \\ 

\hline

\multirow{2}{*}{eating roti} & 2s-AGCN           & 0.90     & 0.75     & 0.99 & 0.83\\
& CTR-GCN &    0.81  &   0.73  & 0.99 & 0.87 \\  
\hline
\end{tabular}
}
\caption{Performance comparison in different categories.}
\label{tab:performance comparison}
\end{table}

Next, we investigate the result in the category of eating roti. Similarly, we tune the confidence threshold $\epsilon$ of \texttt{ProbCost} in the validation set. 
The result is shown in Figure~\ref{fig:TuningThresholdValFood}. It indicates the highest accuracy is achieved at the threshold of 0.1. Therefore, we set the confidence threshold $\epsilon$ of ProbCost as $0.1$ for the evaluation on the testing set.

We compare the performance after \texttt{ProbCost} alignment  with the original output performance from 2s-AGCN and CTR-GCN models. The comparative results are shown in Table~\ref{tab:performance comparison}. After alignment with the process model, both models show improved performance. For example, initially, the 2s-AGCN model achieved an accuracy of 0.90 and a Macro F1 score of 0.75, indicating a strong performance in identifying activities based on the model's predictions. However, after applying the alignment process via \texttt{ProbCost}, there was a notable increase in both metrics. The improvement in the Macro F1 score suggests a better balance between precision and recall across different classes. This enhancement underscores the effectiveness of incorporating process information into the activity recognition. The alignment essentially calibrates the model's predictions to better reflect the actual activity sequences defined by the process model. 

We illustrate an example of activity classes being accurately rectified through the alignment.  Consider an instance where the 2s-AGCN model predicted the sequence of activities as i) \textit{pick food from utensil with both hands}, ii) \textit{pick food from utensil with one hand}, iii) \textit{chewing}, and iv) \textit{move hand towards mouth}. However, the actual sequence, or the ground truth, is i) \textit{pick food from utensil with both hands}, ii) \textit{pick food from utensil with one hand}, iii) \textit{food in hand at table}, and iv) \textit{move hand towards mouth}. This discrepancy indicated that the third activity, \textit{food in hand at table}, was incorrectly identified as \textit{chewing} by the 2s-AGCN model due to the probability distribution (see Table~\ref{tab:prob dist correct} in Appendix), which indicates \textit{chewing} having the highest probability.

Upon incorporating process information through \texttt{ProbCost}, we are able to rectify this misrecognition. The process model of eating roti (Figure~\ref{fig:RotiProcess}) suggests that if the third activity were indeed \textit{chewing}, the subsequent action of \textit{move hand towards mouth} would be illogical. Given the high confidence (0.8 probability of GCNs predictions) in recognising the fourth activity as \textit{move hand towards mouth}, it was clear that if the third activity is \textit{food in hand at table}, the sequence adheres more coherently to the discovered process model. Specifically, the activity of \textit{food in hand at table} indicates that the individual is holding food in their hand while at the table, prepared to bring the food to their mouth for eating. Thus, despite \textit{chewing} initially appearing as the most probable activity for the third step, we identified \textit{food in hand at table} as the correct activity, aligning with the process model. This adjustment showcases the concrete example of how the \texttt{ProbCost} integrates process information with data-driven models to enhance activity recognition accuracy.

\section{Related work}

There are three main categories of research related to HAR: data-driven HAR, knowledge-driven HAR and hybrid HAR that integrates both data and knowledge. 

ML and DL techniques have been extensively applied in exploiting data for activity recognition. The basic idea behind these techniques is extracting meaningful features for activity recognition tasks~\cite{aminikhanghahiEnhancingActivityRecognition2019,deepLeveragingCNNTransfer2019}. It has been demonstrated that DL-based techniques have shown superiority over the other techniques. However, both ML and DL techniques require a large amount of data for training.

Besides the traditional data-driven approaches leveraging large datasets for training models, there is another strand of research focusing on modelling human activity patterns based on prior knowledge, known as knowledge-driven approaches. For example, Chen et al.\ propose an ontological activity modelling and representation approach for activity recognition, such as modelling the interrelations between activities and objects~\citep{chenKnowledgeDrivenApproachActivity2012}.

There are also research initiatives that incorporate domain knowledge into data-driven models for recognising human activities, i.e., hybrid approaches. By integrating domain knowledge, these models can guide the learning process, ensuring that the patterns recognised by the machine learning algorithms align with established theoretical understandings of human behaviour. For example, Asim et al.\ propose an approach for recognising fine-grained activities based on smartphone accelerometer data by incorporating location information~\citep{asimContextAwareHumanActivity2020}.

Our work also incorporates domain knowledge to refine the data-driven models' output. Specifically, we leverage process information from context to enhance the performance of data-driven HAR models.

\section{Conclusion}\label{sec:conclusion HAR}

We present a novel approach aimed at enhancing human activity recognition by integrating process information into the analysis of machine learning model outputs. We apply \texttt{ProbCost} to achieve an alignment between probabilistic event traces generated by ML models and the process model discovered from context. The alignment essentially calibrates the model's predictions by incorporating the actual sequence and nature of activities as defined by the process model. Through experimental comparison based on the Eatsense dataset, with the original performances of Graph Convolutional Networks, including 2s-AGCN and CTR-GCN, our approach demonstrates significant improvement in activity recognition accuracy and Macro F1 scores. 

Our results underscore the effectiveness of incorporating process information for activity recognition. We consider the inherent flexibility of performing activities when discovering process models. Our work also highlights the importance of combining machine learning models with domain-specific process knowledge. Future research can build on these findings by exploring the application of the process model alignment framework to other activities and domains.

\bibliography{refs-condensed}

\clearpage
% \appendices
\appendix
\section{Appendix}
\subsection{Dataset description}

\begin{table}[h]
\centering
\begin{tabular}{cl}
\hline
No.& Activities  \\ \hline
1& chewing                                           \\
2& move hand towards mouth                         \\
3& pick food from utensil with tools in both hands \\
4& pick food from utensil with one hand            \\
5& food in hand at table                           \\
6& pick food from utensil with tool in one hand    \\
7& pick up tools with both hands                   \\
8& move hand away from mouth                       \\
9& no action                                       \\
10& other                                           \\
11& pick up a cup/glass                             \\
12& put the cup/glass back                          \\
13& put one tool back                               \\
14& pick food from utensil with both hands          \\
15& eat it                                          \\
16& drink                                          \\
\hline
\end{tabular}
\caption{The list of eating actions included in the datasets.}
\label{tab:eating actions appendix}
\end{table}

\begin{table}[h]
\centering
\begin{tabular}{cccc}
\hline
Subject & $\#$Videos & Food Type      & $\#$Videos \\ \hline
0       & 5                 & Rice           & 4                 \\
1       & 19                & Toast          & 4                 \\
2       & 2                 & Only-Drinks    & 3                 \\
3       & 2                 & Roti           & 13                \\
4       & 2                 & Egg            & 1                 \\
5       & 2                 & Soup           & 3                 \\
        &                   & Wafers         & 1                 \\
        &                   & Chicken-steaks & 1                 \\
        &                   & Noodles        & 2                 \\ \hline
\end{tabular}
\caption{The number of videos for each subject and food category.}
\label{tab:number of videos appendix}
\end{table}

\subsection{Experiment setting}

Graph Convolutional Networks we used in Section~\ref{sec:experiment HAR}, including 2s-AGCN and CTR-GCN are implemented in mmaction2 by opemmmlab~\cite{2020mmaction2}. The hyperparameters are specified in Table~\ref{tab:hyperparameter}. We use Stochastic Gradient Descent (SGD) with momentum 0.9 as the optimiser of GCNs. 

\begin{table}[ht]
\centering
\begin{tabular}{c|c|c}
\hline
GCN Model                    & Hypterparameter & Value        \\ \hline
\multirow{5}{*}{2s-AGCN} & learning rate     & 0.1          \\
                         & weight\_decay  & 0.0001       \\
                         & loss            & Cross-entropy \\
                         & batch\_size     & 16           \\
                         & momentum        & 0.9          \\ \hline
\multirow{5}{*}{CTR-GCN} & learning rate    & 0.1             \\
                         & weight\_decay    &  0.0001       \\ 
                         & loss            & Cross-entropy \\
                         & batch\_size     & 16           \\
                         & momentum        & 0.9          \\
                         
                         \hline
\end{tabular}
\caption{The hyperparameters of GCN models.}
\label{tab:hyperparameter}
\end{table}

\vspace{5cm}

\subsection{Experiment results}

\begin{table}[ht]
\centering
\begin{small}

% \resizebox{\columnwidth}{!}{
\begin{tabular}{lc}
\hline
Activities & Probabilities \\ \hline
chewing                                         & 0.352730  \\
move hand towards mouth                         & 0.325867 \\
pick food from utensil with tools in both hands & 0.082761 \\
pick food from utensil with one hand            & 0.082365 \\
food in hand at table                           & 0.034219 \\
pick food from utensil with tool in one hand    & 0.029754 \\
pick up tools with both hands                   & 0.026519 \\
move hand away from mouth                       & 0.024972 \\
no action                                       & 0.009012 \\
other                                           & 0.008487 \\
pick up a cup/glass                             & 0.007892 \\
put the cup/glass back                          & 0.006343 \\
put one tool back                               & 0.005486 \\
pick food from utensil with both hands          & 0.002836 \\
eat it                                          & 0        \\
drink                                           & 0       \\
\hline
\end{tabular}
% }
\end{small}
\caption{The probability distribution of an activity, which is corrected after alignment.}
\label{tab:prob dist correct}
\end{table}

% \clearpage

% \vspace{1cm}

\end{document}